\documentclass[11pt]{article}
\usepackage{coling2020}
\usepackage{times}
\usepackage{url}
\usepackage{latexsym}

\usepackage{todonotes}
\usepackage{subcaption}
\usepackage{tabularx} 
\usepackage{diagbox}
\usepackage{graphicx}
\usepackage{datetime}
\usepackage{graphicx}
\usepackage{caption}
\usepackage{subcaption}
\usepackage{pgfplots}
\usepackage{pgfplotstable}
\pgfplotsset{compat=1.7}
\usepackage{tikz}
\usepackage{paralist}
\usepackage{booktabs}
\usepackage{multirow}
\usepackage{mathtools}
\usepackage{xcolor}
\usepackage{hyperref}

\colingfinalcopy 

\title{The Devil is in the Details: Evaluating Limitations of Transformer-based Methods for Granular Tasks}

\author{
 Brihi Joshi$^{\dagger}$ ~ Neil Shah$^{\clubsuit}$ ~ Francesco Barbieri$^{\clubsuit}$ ~ Leonardo Neves$^{\clubsuit}$  \vspace{0.1cm} \\
$^\dagger$Indraprastha Institute of Information Technology Delhi, New Delhi, India \vspace{0.1cm} \\
$^\clubsuit$Snap Inc., Santa Monica, CA 90405, USA\\
{  \tt $^{\dagger}$brihi16142@iiitd.ac.in},\\ 
{ \tt $^{\clubsuit}$\{nshah,fbarbieri,lneves\}@snap.com},\\
}

\date{}

\begin{document}
\maketitle
\begin{abstract}

Contextual embeddings derived from transformer-based neural language models have shown state-of-the-art performance for various tasks such as question answering, sentiment analysis, and textual similarity in recent years. Extensive work shows how accurately such models can represent \emph{abstract}, semantic information present in text. In this expository work, we explore a tangent direction and analyze such models' performance on tasks that require a more \emph{granular} level of representation.  We focus on the problem of textual similarity from two perspectives: matching documents on a granular level (requiring embeddings to capture fine-grained attributes in the text), and an abstract level (requiring  embeddings to capture overall textual semantics). We empirically demonstrate, across two 
datasets from different domains, that despite high performance in abstract document matching as expected, contextual embeddings are consistently (and at times, vastly) outperformed by simple baselines like TF-IDF for more granular tasks. We then propose a simple but effective method to incorporate TF-IDF into models that use contextual embeddings, achieving relative improvements of up to 36\% on granular tasks.
\end{abstract}

\section{Introduction}





In recent years, contextual embeddings \cite{peters-etal-2018-deep,devlin2018bert} have made immense progress in semantic understanding-based tasks.  After being trained using large amounts of data, for example via a self-supervised task like masked language-modeling, such models learn crucial elements of language, such as syntax and semantics \cite{jawahar2019does,goldberg2019assessing,wiedemann2019does} from just raw text.
The best performing contextual embeddings are trained with Transformer-based methods (TBM) \cite{vaswani2017attention,devlin2018bert}. These embeddings have been shown to frequently achieve state-of-the-art results in downstream tasks like question answering and sentiment analysis \cite{bert-qa,sun-etal-2019-utilizing}.
Contextual embeddings are also often used to capture the similarity between pairs of documents; for example, on the Semantic Textual Similarity (STS) task \cite{cer-etal-2017-semeval}  included in the GLUE benchmark \cite{wang2018glue}, TBMs have shown competitive performance, substantially outperforming embedding baselines like Word2Vec \cite{word2vec} and GloVE \cite{Pennington14glove}. However, their performance on similarity tasks beyond abstract, semantic ones \cite{mickus2019mean} 
--  for example, on granular news article matching --  is less understood.

\begin{figure}[ht]
\centering
  \includegraphics[width=0.9\linewidth]{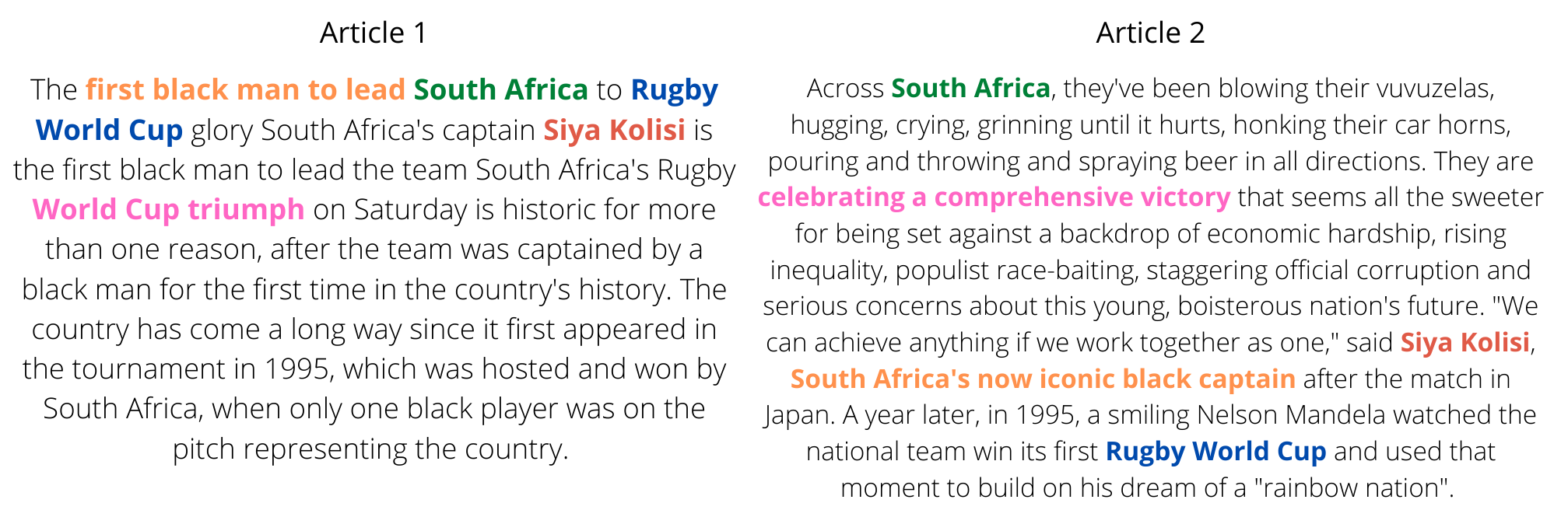}  
  \caption{An example pair of articles from the \textbf{News Dedup} dataset: Both report the same news event, and are thus \emph{similar on a granular level}; the colored text indicates fine-grained details associated with this determination.  Both articles are also of the ``sports'' topic, and are thus \emph{similar on an abstract level}.
  }
  \label{fig:example_articles}
\end{figure}

In this work, we study the performance of TBMs in textual similarity tasks with the following research question:  
\emph{Are transformer-based methods as performant for granular tasks as they are for abstract ones?} Here, \emph{granular} and \emph{abstract} reflect varying amounts of coarseness in the concept of \emph{similarity}. For example, consider the news domain: A granular notion of similarity might be whether a pair of articles both report the exact same news event.  Conversely, an abstract notion might be when the articles share the same topical category, like sports or finance. Figure \ref{fig:example_articles} illustrates this with an example for clarity.

Firstly, we define separate tasks to explore these two notions of similarity on two datasets from different domains -- News Articles, and Bug Reports. Our analysis on both datasets reveals that contextual embeddings \emph{do not} perform well on granular tasks, and are outperformed by simple baselines like TF-IDF. Secondly, we demonstrate that TBM contextual embeddings 
\emph{do} in fact contain important semantic information, and a simple interpolation strategy between the two methods can help boost the relative individual performance of TBMs (TF-IDF) by up to 36\% (6\%) 
on the granular task.


\section{Related Work}
We discuss related work in two areas: textual similarity, and TBMs.

\textbf{Textual Similarity} has been studied from various perspectives -- comparing documents of different lengths in order to capture varying levels of detail \cite{gong-etal-2018-document}, evaluating semantic similarity between reference and generated corpus \cite{clark-etal-2019-sentence}, and semantic similarity for long documents in a hierarchical fashion \cite{smash-rnn}. It is also shown that sentence meta-embeddings (obtained from combining ensembles of sentence embeddings) perform better \cite{poerner2019sentence} for semantic similarity tasks compared to the individual baselines. For duplicate detection, which is a more granular task compared to semantic similarity, \newcite{rodier-carter-2020-online} show that detection of near-duplicates in news articles can be identified by evaluating $n$-gram level overlap in documents. In the news domain, \newcite{liu2018matching} shows that article similarity can be improved by extracting common `concepts' from the two articles using graph-based approaches. 

\textbf{TBMs} \cite{liu2019roberta,devlin2018bert} have been shown to consistently perform better in the semantic similarity tasks. \newcite{peinelt-etal-2020-tbert} also shows that BERT-based architectures appended with topic-related details from topic models lead to an increase in semantic similarity performance. However, few works have highlighted TBMs' ability to capture granular information. \cite{khattab2020colbert} shows that BERT can be used for document retrieval by matching embeddings of each word in the query and document, capturing granular similarity.

Unlike previous approaches that focus on either a granular or abstract similarity task, we compare the performance of TBMs with other baseline methods across the two tasks, and in addition, provide a simple method to improve the performance of TBMs on granular similarity tasks.

\section{Method}
\label{method}

In this section, we describe the methodology used to compare two documents from a granular and abstract perspective. Further, we also define the granular and abstract text similarity tasks in detail.

\subsection{Problem Definition}

We consider both granular and abstract tasks to be similarity classification tasks operating on a pair of documents.  The task-specific labels are binary, indicating whether the pair is judged to be similar or not (one label for abstract, one for granular).  From a corpus $\mathcal{C}$, we consider a pair of two documents $d_{1}, d_{2}$  and their task-specific similarity judgment $y$ (without loss of generality).  We define $e_{k} = f(d_{k})$, where $e_{k}$ is $d_{k}$'s embedding, produced by $f(\cdot)$. In practice, $f$ could be a vector space method like TF-IDF, or the final layer from a TBM. Upon obtaining $e_{1}, e_{2}$, we generate a (symmetric) pair similarity score $g(e_{1}, e_{2})$ corresponding to the given task, and use it to arrive at a binary prediction $\hat{y}$.  Performance is measured using standard metrics quantifying agreement between $\hat{y}$ and $y$ across pairs.

\begin{figure}[ht]
\centering
  \includegraphics[width=0.9\linewidth]{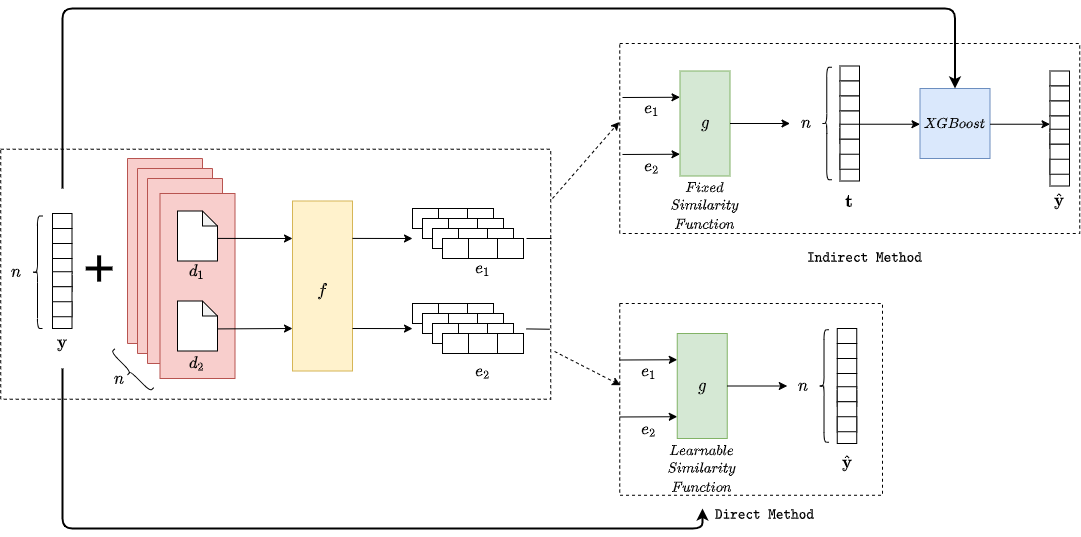}  
  \caption{Our experimentation setups take $n$ pairs consisting of two documents ($d_1$, $d_2$) and their similarity label ($y$) and yields their similarity score ($\hat{y}$)
  }
  \label{fig:flowchart}
\end{figure}

\subsection{Experimental Setup}
We consider two experimental settings: \emph{indirect} and \emph{direct}. These are also illustrated in Figure \ref{fig:flowchart}\footnote{The code for our experiments is available at \href{https://github.com/brihijoshi/granular-similarity-COLING-2020}{ https://github.com/brihijoshi/granular-similarity-COLING-2020}}. In the \emph{indirect} setting, we indirectly learn to predict $\mathbf{y}$ via a fixed $g$. Specifically, we a priori define $g$ as the cosine similarity function, and define a vector $\mathbf{t}$, with each entry corresponding to $g(e_1, e_2)$ for a given pair of document embeddings w.l.o.g. We then feed $\mathbf{t}$ as a feature and the task-specific label vector $\mathbf{y}$ to XGBoost \cite{Chen_2016} to obtain the predictions $\mathbf{\hat{y}}$.  We evaluate using several embedding functions:
\begin{compactitem}
    \item \textbf{TF-IDF:} TF-IDF \cite{ramos2003using} weights corresponding to the $1$-gram tokens inserted in their respective indices in an array of same length as the train's set vocabulary.
    \item \textbf{WME:} Word Mover's Embedding (WME) \cite{wu-etal-2018-word} generated from static embeddings like word2vec \cite{word2vec} using the Word Mover's Distance metric \cite{wmd}.
    \item \textbf{SIF:} The SIF\cite{arora2017asimple} weighting scheme employed over pretrained GloVE embeddings.
    \item \textbf{RT:} The embedding corresponding to the \texttt{CLS} token in the final layer of a pretrained RoBERTa \cite{liu2019roberta} model. 
    \item \textbf{LF:} The embedding corresponding to the \texttt{CLS} token in the final layer of a pretrained LongFormer \cite{Beltagy2020Longformer} model. 
    \item \textbf{ST-RT:} Sentence embeddings generated using Sentence Transformers \cite{reimers-2019-sentence-bert}, which is a RoBERTa model fine-tuned on the STS benchmark.
\end{compactitem}


In the \emph{direct} setting, we  directly learn to predict $\mathbf{y}$ from embeddings $e_1, e_2$ in an end-to-end manner, rather than through a predefined similarity measure. We use the best performing embeddings from the previous setting as input to train the model $g$ which produces a score, which is thresholded to derive $\mathbf{\hat{y}}$.


\begin{compactitem}
    \item \textbf{TF-IDF-E2E:} We compute the absolute difference between the TF-IDF vectors of the article pairs and train a Logistic Regression classifier with the labels corresponding to the similarity task.
    
 \item \textbf{RT-E2E:} Since ST-RT embedding uses a pre-trained RoBERTa model, we train RoBERTa end-to-end instead (as \newcite{reimers-2019-sentence-bert} mentions, ST is not intended for end-to-end use). We provide the article pairs to a pre-trained RoBERTa model, separated with the \texttt{SEP} token. It is then directly fine-tuned on the task-specific labels.

\end{compactitem}

\subsection{Datasets}

\begin{table}[t!]
\centering
\scalebox{.85}{
\begin{tabular}{cccccccc}
\toprule
 & \multicolumn{3}{c}{\textbf{Whole dataset}} & \multicolumn{2}{c}{\textbf{Granular}} & \multicolumn{2}{c}{\textbf{Abstract}} \\
 
  \cmidrule(lr){2-4} \cmidrule(lr){5-6} \cmidrule(lr){7-8}
\textbf{} &
  \textbf{\begin{tabular}[c]{@{}c@{}}Avg \# words\end{tabular}} &
  \textbf{\begin{tabular}[c]{@{}c@{}}Train\end{tabular}} &
  \textbf{\begin{tabular}[c]{@{}c@{}}Test\end{tabular}} &
  \textbf{\begin{tabular}[c]{@{}c@{}}Train\end{tabular}} &
  \textbf{\begin{tabular}[c]{@{}c@{}}Test\end{tabular}} &
  \textbf{\begin{tabular}[c]{@{}c@{}}Train\end{tabular}} &
  \textbf{\begin{tabular}[c]{@{}c@{}}Test\end{tabular}} \\ \hline
\textbf{ND} &
  388.6 &
  3105 &
  695 &
  1763/1342 &
  53/642 &
  2452/653 &
  509/186 \\ 
\textbf{BR} &
  5.7 &
  72142 &
  8220 &
  40967/31175 &
  3954/4266 &
  47199/24943 &
  4804/3416 \\ 
  \bottomrule
\end{tabular}
}
\caption{\label{tab:dataset_details} Dataset and evaluation split details: The dataset statistics capture number of unique \emph{text pairs}. For the task-specific statistics, we report the total number of similar pairs and non-similar (similar/not-similar) pairs according to the task for each split. The imbalance in the test set of \textbf{ND} replicates the distribution found in the real-world news event similarity detection problems.}
\end{table}

We evaluate with datasets from News Articles 
and Bug Reports domains to demonstrate generality. Each includes both abstract and granular labels for the same documents.

\textbf{News Dedup dataset (ND)} contains pairs of news articles from 243 different English news sources, collected between September and November 2019 from RSS feeds.  For each pair, we assign a granular binary label indicating whether the pair reports the same news event, and an abstract binary label reflecting whether they share the same topic (politics, business, technology, entertainment, sports, science, or other -- adapted from Google News\footnote{https://news.google.com/}). We source annotations from Amazon Mechanical Turk\footnote{https://www.mturk.com/}, relying on multiple annotator agreement, with Fleiss' $\kappa$ coefficient of $0.68$. See Appendix A for details.




\textbf{Bug Repo dataset (BR)} \cite{bugrepo} contains bug reports from several open-source projects like Eclipse and Mozilla, and is used primarily for duplicate bug detection. Each report consists of a title, a description of the error, the broad category that the bug belongs to (e.g. UI or Scripting out of $21$ others), and a set of duplicate reports that flag the same bug.  We indicate granular similarity as those pairs which flag the same bug, and abstract similarity as those pairs which fall under the same category, with the title of the report as the textual input.
For each dataset, the documents in the train and test splits are disjoint sets, ensuring that the model does not memorize textual representations. For ND, the sets are also temporally disjoint, avoiding event overlap between train and test splits. Further details about the splits are provided in Table \ref{tab:dataset_details}.

\section{Results} 




\begin{table}[t!]
\centering
\setlength{\tabcolsep}{5.5pt}
\scalebox{.85}{
\begin{tabular}{cccccccccc}
\toprule
 & & \multicolumn{6}{c}{\textbf{Indirect (Cosine Sim.)}} & \multicolumn{2}{c}{\textbf{Direct (End-to-end)}} \\ \cmidrule(lr){3-8} \cmidrule(lr){9-10}
 & & \textbf{TF-IDF} & \textbf{WMD} & \textbf{SIF} & \textbf{RT} & \textbf{LF} & \textbf{ST-RT} & \textbf{TF-IDF-E2E} & \textbf{RT-E2E} \\ \hline
\multirow{2}{*}{\textbf{Granular}} & 
\textbf{ND} & \textbf{0.85} & 0.72 & 0.54 & 0.59 & 0.62 & 0.66 & 0.68 & 0.59 \\ 
 & \textbf{BR} & \textbf{0.75} & 0.62 & 0.43 & 0.66 & 0.69 & 0.71 & 0.71 & 0.70 \\ \hline 
 
\multirow{2}{*}{\textbf{Abstract}} & 
\textbf{ND} & 0.54 & 0.57 & 0.51 & 0.59 & 0.62 & 0.62 & 0.58 & \textbf{0.66} \\ 
 & \textbf{BR} & 0.69 & 0.47 & 0.40 & 0.67 & 0.70 & 0.73 & 0.51 & \textbf{0.74} \\ 
 \bottomrule
\end{tabular}
}
\caption{\label{tab:abstract} Granular and abstract similarity results: TF-IDF outperforms TBMs on granular tasks, while TBMs outperform on abstract tasks in both settings.}
\label{tab:results}
\end{table}

\begin{table}[tb!]
\centering
\setlength{\tabcolsep}{5.5pt}
\scalebox{.85}{
\begin{tabular}{ccccccccc}
 \toprule
 &  & \multicolumn{7}{c}{\textbf{Values of weight $w$}} \\
 \cmidrule(lr){3-9}
 &  & \textbf{0} & \textbf{0.1} & \textbf{0.3} & \textbf{0.5} & \textbf{0.7} & \textbf{0.9} & \textbf{1} \\ 
 \hline
\multirow{2}{*}{\textbf{Granular}} 
& \textbf{ND} & 0.66 & 0.77 & 0.83 & 0.89 & \textbf{0.90} & 0.86 & 0.85 \\
 & \textbf{BR} & 0.71 & 0.56 & 0.62 & 0.69 & \textbf{0.79} & 0.76 & 0.75 \\
 \hline
\multirow{2}{*}{\textbf{Abstract}} 
& \textbf{ND} & \textbf{0.62} & 0.60 & 0.55 & 0.56 & 0.60 & 0.57 & 0.54 \\
 & \textbf{BR} & \textbf{0.73} & 0.72 & 0.72 & 0.70 & 0.69 & 0.68 & 0.69 \\
\bottomrule
\end{tabular}
}
\caption{\label{tbl:merge_results} Performance for granular and abstract tasks 
the 2 datasets as we vary the value of \textit{w}. Note that best granular results are achieved by interpolating TF-IDF with TBM predictions ($w=0.7$).}
\end{table}

Table \ref{tab:results} summarizes the experiments on the two
datasets using the methods mentioned in Section \ref{method}. We can observe that a simple TF-IDF based approach performs better than all embedding methods for the granular-level similarity tasks.
However, for the abstract-level similarity task, training a RoBERTa model to perform the task end-to-end achieves, as expected, the highest performance.

Despite the better results, the complete absence of semantic understanding is a disadvantage of TF-IDF. To mitigate this issue, we propose a simple approach to merge the best performing indirect methods. Let $g_{t}$ and $g_{r}$ be the similarity scores obtained from the TF-IDF and the ST-RT approaches respectively: we obtain a new, interpolated score $g_{i} = w \cdot g_{t} + (1-w) \cdot g_{r}$, that is then used as an input to the classifier. As Table 3 shows,  performance drastically changes when varying $w$. For both datasets, we observe the best results when $w=0.7$, demonstrating that combining semantic and fine-grained information is helpful for granular tasks. \footnote{Even though $w$ is robust to the two datasets that we are using, we would recommend tuning it for other datasets.}
Conversely, we achieve the best performance on the abstract level when using only ST-RT. We hypothesize the noise introduced by the granular information results in the performance drop in cases prioritizing abstract, semantic relevance.



\section{Conclusion}
In this work, we study the use of contextual embeddings derived from transformer-based models (TBMs) for semantic similarity tasks of varying granularity level. Through empirical analysis, we show that while TBMs achieve higher performance in the abstract similarity tasks, simple methods like TF-IDF outperform these models for granular similarity tasks (like event matching). We then propose a simple but effective method to merge these two approaches, achieving relative improvements of 36\% (6\%) when compared to using only TBMs (TF-IDF).  In future work, we plan to investigate the scope for integrating granular information into TBM contextual embeddings to toggle the granularity that such embeddings inherently encode.

\bibliographystyle{coling}
\bibliography{coling2020}

\begin{thebibliography}{}

\bibitem[\protect\citename{Arora \bgroup et al.\egroup }2017]{arora2017asimple}
Sanjeev Arora, Yingyu Liang, and Tengyu Ma.
\newblock 2017.
\newblock A simple but tough-to-beat baseline for sentence embeddings.

\bibitem[\protect\citename{Beltagy \bgroup et al.\egroup
  }2020]{Beltagy2020Longformer}
Iz~Beltagy, Matthew~E. Peters, and Arman Cohan.
\newblock 2020.
\newblock Longformer: The long-document transformer.
\newblock {\em arXiv:2004.05150}.

\bibitem[\protect\citename{Cer \bgroup et al.\egroup
  }2017]{cer-etal-2017-semeval}
Daniel Cer, Mona Diab, Eneko Agirre, I{\~n}igo Lopez-Gazpio, and Lucia Specia.
\newblock 2017.
\newblock {S}em{E}val-2017 task 1: Semantic textual similarity multilingual and
  crosslingual focused evaluation.
\newblock In {\em Proceedings of the 11th International Workshop on Semantic
  Evaluation ({S}em{E}val-2017)}, pages 1--14, Vancouver, Canada, August.
  Association for Computational Linguistics.

\bibitem[\protect\citename{Chen and Guestrin}2016]{Chen_2016}
Tianqi Chen and Carlos Guestrin.
\newblock 2016.
\newblock Xgboost.
\newblock {\em Proceedings of the 22nd ACM SIGKDD International Conference on
  Knowledge Discovery and Data Mining}, Aug.

\bibitem[\protect\citename{Clark \bgroup et al.\egroup
  }2019]{clark-etal-2019-sentence}
Elizabeth Clark, Asli Celikyilmaz, and Noah~A. Smith.
\newblock 2019.
\newblock Sentence mover{'}s similarity: Automatic evaluation for
  multi-sentence texts.
\newblock In {\em Proceedings of the 57th Annual Meeting of the Association for
  Computational Linguistics}, pages 2748--2760, Florence, Italy, July.
  Association for Computational Linguistics.

\bibitem[\protect\citename{Devlin \bgroup et al.\egroup }2018]{devlin2018bert}
Jacob Devlin, Ming-Wei Chang, Kenton Lee, and Kristina Toutanova.
\newblock 2018.
\newblock Bert: Pre-training of deep bidirectional transformers for language
  understanding.

\bibitem[\protect\citename{Goldberg}2019]{goldberg2019assessing}
Yoav Goldberg.
\newblock 2019.
\newblock Assessing bert's syntactic abilities.
\newblock {\em arXiv preprint arXiv:1901.05287}.

\bibitem[\protect\citename{Gong \bgroup et al.\egroup
  }2018]{gong-etal-2018-document}
Hongyu Gong, Tarek Sakakini, Suma Bhat, and JinJun Xiong.
\newblock 2018.
\newblock Document similarity for texts of varying lengths via hidden topics.
\newblock In {\em Proceedings of the 56th Annual Meeting of the Association for
  Computational Linguistics (Volume 1: Long Papers)}, pages 2341--2351,
  Melbourne, Australia, July. Association for Computational Linguistics.

\bibitem[\protect\citename{Hamborg \bgroup et al.\egroup }2017]{Hamborg2017}
Felix Hamborg, Norman Meuschke, Corinna Breitinger, and Bela Gipp.
\newblock 2017.
\newblock news-please: A generic news crawler and extractor.
\newblock In Maria Gaede, Violeta Trkulja, and Vivien Petra, editors, {\em
  Proceedings of the 15th International Symposium of Information Science},
  pages 218--223, March.

\bibitem[\protect\citename{Jawahar \bgroup et al.\egroup
  }2019]{jawahar2019does}
Ganesh Jawahar, Beno{\^\i}t Sagot, and Djam{\'e} Seddah.
\newblock 2019.
\newblock What does bert learn about the structure of language?
\newblock In {\em Proceedings of the 57th Annual Meeting of the Association for
  Computational Linguistics}.

\bibitem[\protect\citename{Jiang \bgroup et al.\egroup }2019]{smash-rnn}
Jyun-Yu Jiang, Mingyang Zhang, Cheng Li, Michael Bendersky, Nadav Golbandi, and
  Marc Najork.
\newblock 2019.
\newblock Semantic text matching for long-form documents.
\newblock In {\em The World Wide Web Conference}, WWW ’19, page 795–806,
  New York, NY, USA. Association for Computing Machinery.

\bibitem[\protect\citename{Khattab and Zaharia}2020]{khattab2020colbert}
Omar Khattab and Matei Zaharia.
\newblock 2020.
\newblock Colbert: Efficient and effective passage search via contextualized
  late interaction over bert.

\bibitem[\protect\citename{Kusner \bgroup et al.\egroup }2015]{wmd}
Matt~J. Kusner, Yu~Sun, Nicholas~I. Kolkin, and Kilian~Q. Weinberger.
\newblock 2015.
\newblock From word embeddings to document distances.
\newblock In {\em Proceedings of the 32nd International Conference on
  International Conference on Machine Learning - Volume 37}, ICML’15, page
  957–966. JMLR.org.

\bibitem[\protect\citename{{Lamkanfi} \bgroup et al.\egroup }2013]{bugrepo}
A.~{Lamkanfi}, J.~{Pérez}, and S.~{Demeyer}.
\newblock 2013.
\newblock The eclipse and mozilla defect tracking dataset: A genuine dataset
  for mining bug information.
\newblock In {\em 2013 10th Working Conference on Mining Software Repositories
  (MSR)}, pages 203--206.

\bibitem[\protect\citename{Liu \bgroup et al.\egroup }2018]{liu2018matching}
Bang Liu, Di~Niu, Haojie Wei, Jinghong Lin, Yancheng He, Kunfeng Lai, and
  Yu~Xu.
\newblock 2018.
\newblock Matching article pairs with graphical decomposition and convolutions.

\bibitem[\protect\citename{Liu \bgroup et al.\egroup }2019]{liu2019roberta}
Yinhan Liu, Myle Ott, Naman Goyal, Jingfei Du, Mandar Joshi, Danqi Chen, Omer
  Levy, Mike Lewis, Luke Zettlemoyer, and Veselin Stoyanov.
\newblock 2019.
\newblock Roberta: A robustly optimized bert pretraining approach.

\bibitem[\protect\citename{Mickus \bgroup et al.\egroup }2019]{mickus2019mean}
Timothee Mickus, Denis Paperno, Mathieu Constant, and Kees van Deemter.
\newblock 2019.
\newblock What do you mean, bert? assessing bert as a distributional semantics
  model.

\bibitem[\protect\citename{Mihalcea and
  Tarau}2004]{mihalcea-tarau-2004-textrank}
Rada Mihalcea and Paul Tarau.
\newblock 2004.
\newblock {T}ext{R}ank: Bringing order into text.
\newblock In {\em Proceedings of the 2004 Conference on Empirical Methods in
  Natural Language Processing}, pages 404--411, Barcelona, Spain, July.
  Association for Computational Linguistics.

\bibitem[\protect\citename{Mikolov \bgroup et al.\egroup }2013]{word2vec}
Tomas Mikolov, Ilya Sutskever, Kai Chen, Greg Corrado, and Jeffrey Dean.
\newblock 2013.
\newblock Distributed representations of words and phrases and their
  compositionality.
\newblock In {\em Proceedings of the 26th International Conference on Neural
  Information Processing Systems - Volume 2}, NIPS’13, page 3111–3119, Red
  Hook, NY, USA. Curran Associates Inc.

\bibitem[\protect\citename{Peinelt \bgroup et al.\egroup
  }2020]{peinelt-etal-2020-tbert}
Nicole Peinelt, Dong Nguyen, and Maria Liakata.
\newblock 2020.
\newblock t{BERT}: Topic models and {BERT} joining forces for semantic
  similarity detection.
\newblock In {\em Proceedings of the 58th Annual Meeting of the Association for
  Computational Linguistics}, pages 7047--7055, Online, July. Association for
  Computational Linguistics.

\bibitem[\protect\citename{Pennington \bgroup et al.\egroup
  }2014]{Pennington14glove}
Jeffrey Pennington, Richard Socher, and Christopher~D. Manning.
\newblock 2014.
\newblock Glove: Global vectors for word representation.
\newblock In {\em In EMNLP}.

\bibitem[\protect\citename{Peters \bgroup et al.\egroup
  }2018]{peters-etal-2018-deep}
Matthew Peters, Mark Neumann, Mohit Iyyer, Matt Gardner, Christopher Clark,
  Kenton Lee, and Luke Zettlemoyer.
\newblock 2018.
\newblock Deep contextualized word representations.
\newblock In {\em Proceedings of the 2018 Conference of the North {A}merican
  Chapter of the Association for Computational Linguistics: Human Language
  Technologies, Volume 1 (Long Papers)}, pages 2227--2237, New Orleans,
  Louisiana, June. Association for Computational Linguistics.

\bibitem[\protect\citename{Poerner \bgroup et al.\egroup
  }2019]{poerner2019sentence}
Nina Poerner, Ulli Waltinger, and Hinrich Schütze.
\newblock 2019.
\newblock Sentence meta-embeddings for unsupervised semantic textual
  similarity.

\bibitem[\protect\citename{Ramos and others}2003]{ramos2003using}
Juan Ramos et~al.
\newblock 2003.
\newblock Using tf-idf to determine word relevance in document queries.
\newblock In {\em Proceedings of the first instructional conference on machine
  learning}, volume 242, pages 133--142. New Jersey, USA.

\bibitem[\protect\citename{Reimers and
  Gurevych}2019]{reimers-2019-sentence-bert}
Nils Reimers and Iryna Gurevych.
\newblock 2019.
\newblock Sentence-bert: Sentence embeddings using siamese bert-networks.
\newblock In {\em Proceedings of the 2019 Conference on Empirical Methods in
  Natural Language Processing}. Association for Computational Linguistics, 11.

\bibitem[\protect\citename{Rodier and Carter}2020]{rodier-carter-2020-online}
Simon Rodier and Dave Carter.
\newblock 2020.
\newblock Online near-duplicate detection of news articles.
\newblock In {\em Proceedings of The 12th Language Resources and Evaluation
  Conference}, pages 1242--1249, Marseille, France, May. European Language
  Resources Association.

\bibitem[\protect\citename{Sun \bgroup et al.\egroup
  }2019]{sun-etal-2019-utilizing}
Chi Sun, Luyao Huang, and Xipeng Qiu.
\newblock 2019.
\newblock Utilizing {BERT} for aspect-based sentiment analysis via constructing
  auxiliary sentence.
\newblock In {\em Proceedings of the 2019 Conference of the North {A}merican
  Chapter of the Association for Computational Linguistics: Human Language
  Technologies, Volume 1 (Long and Short Papers)}, pages 380--385, Minneapolis,
  Minnesota, June. Association for Computational Linguistics.

\bibitem[\protect\citename{van Aken \bgroup et al.\egroup }2019]{bert-qa}
Betty van Aken, Benjamin Winter, Alexander L\"{o}ser, and Felix~A. Gers.
\newblock 2019.
\newblock How does bert answer questions? a layer-wise analysis of transformer
  representations.
\newblock In {\em Proceedings of the 28th ACM International Conference on
  Information and Knowledge Management}, CIKM ’19, page 1823–1832, New
  York, NY, USA. Association for Computing Machinery.

\bibitem[\protect\citename{Vaswani \bgroup et al.\egroup
  }2017]{vaswani2017attention}
Ashish Vaswani, Noam Shazeer, Niki Parmar, Jakob Uszkoreit, Llion Jones,
  Aidan~N Gomez, {\L}ukasz Kaiser, and Illia Polosukhin.
\newblock 2017.
\newblock Attention is all you need.
\newblock In {\em Advances in neural information processing systems}, pages
  5998--6008.

\bibitem[\protect\citename{Wang \bgroup et al.\egroup }2018]{wang2018glue}
Alex Wang, Amanpreet Singh, Julian Michael, Felix Hill, Omer Levy, and
  Samuel~R. Bowman.
\newblock 2018.
\newblock Glue: A multi-task benchmark and analysis platform for natural
  language understanding.

\bibitem[\protect\citename{Wiedemann \bgroup et al.\egroup
  }2019]{wiedemann2019does}
Gregor Wiedemann, Steffen Remus, Avi Chawla, and Chris Biemann.
\newblock 2019.
\newblock Does bert make any sense? interpretable word sense disambiguation
  with contextualized embeddings.

\bibitem[\protect\citename{Wu \bgroup et al.\egroup }2018]{wu-etal-2018-word}
Lingfei Wu, Ian En-Hsu Yen, Kun Xu, Fangli Xu, Avinash Balakrishnan, Pin-Yu
  Chen, Pradeep Ravikumar, and Michael~J. Witbrock.
\newblock 2018.
\newblock Word mover{'}s embedding: From {W}ord2{V}ec to document embedding.
\newblock In {\em Proceedings of the 2018 Conference on Empirical Methods in
  Natural Language Processing}, pages 4524--4534, Brussels, Belgium,
  October-November. Association for Computational Linguistics.

\end{thebibliography}

\newpage

\section*{Appendix A - News Dedup (ND) dataset details}

The News Dedup dataset was collected from a set of $243$ English news sources, over a span of 3 months (September 2019 - November 2019). Further details about the dataset are as follows - 
\begin{enumerate}

\item \textbf{Collection: } With a list of pre-prepared news sources, we extract the links to the RSS feeds of these sources. We then use the News-please package \cite{Hamborg2017} to extract details like the title, body, attached images, etc from the updated articles from the RSS feeds. This process is scheduled ever $3$ hours, so that new articles are immediately scraped. Given that most of the news sources report articles belonging to the topic `Politics', we attempted to ensure articles from diverse topics to be equally represented during annotation.

\item \textbf{Postprocessing: } In order to create relevant article pairs from the set of scraped news articles, we first extract keywords using TextRank \cite{mihalcea-tarau-2004-textrank}. We then extract embeddings for the keywords using Word2vec and average the embeddings to obtain an embedding for the entire article. We then use cosine similarity between all possible embeddings in the corpus. This similarity score is used as a `proxy' to extract relevant pairs for annotation. We bin our obtained pairs into 3 categories -- positives, easy negatives and hard negatives. Easy negatives are pairs with a similarity score less than a certain threshold (upon observation, this threshold seemed to be ranging from 20-30\%). They are however, verified during annotation process. Hard negatives are pairs with high similarity, but do not satisfy our criterion of similarity (this happens mostly for the granular task). Before the annotation, we ensured that the dataset did not contain transitive pairs, i.e, if articles A and B and articles B and C are present as pairs, we ensured that articles A and C are not present in the dataset while it is being annotated.

\item \textbf{Annotation: } During the annotation process, each article pair is labelled by $3$ annotators. While constructing the task on Amazon Mechanical Task, it is ensured that the task contains \emph{golden questions} -- a set of common sense questions based on current affairs, so that the annotators are judged on their capability to read lengthy news articles. The annotators had substantial agreement for the Granular simlilarity task (with Fleiss' kappa coefficient of $0.68$).

\end{enumerate}

\end{document}